\documentclass[journal]{IEEEtran}
\usepackage[utf8]{inputenc}
\usepackage[legalpaper, margin=2cm]{geometry}
\usepackage{array, multirow, graphicx }
\usepackage[sorting=none]{biblatex}
\usepackage{authblk}
\bibliography{bibliography}

\newcommand{\beginsupplement}{%
        \setcounter{table}{0}
        \renewcommand{\thetable}{S\arabic{table}}%
        \setcounter{figure}{0}
        \renewcommand{\thefigure}{S\arabic{figure}}%
     }

\ifCLASSOPTIONcompsoc
\usepackage[caption=false,font=normalsize,labelfon
t=sf,textfont=sf]{subfig}
\else
\usepackage[caption=false,font=footnotesize]{subfig}
\fi

\newcolumntype{x}[1]{>{\centering\arraybackslash}m{#1cm}}
\newcommand\tableMarginBottom{ 0.25cm }
\newcommand\tableHline{ \\[-0.21cm] \hline \\[-0.2cm] }
\graphicspath{ { figures/ } }
\DeclareGraphicsExtensions{.pdf,.png}

\title{Population Gradients improve performance across data-sets and architectures in object classification}
\author[1]{ Yurika Sakai } 
\author[2,3]{ Andrey Kormilitzin } 
\author[2,4]{ Qiang Liu } 
\author[2,4]{ Alejo Nevado-Holgado }
\affil[1]{Centre for Tropical Medicine and Global Health, University of Oxford}
\affil[2]{Department of Psychiatry, University of Oxford}
\affil[3]{Institute of Mathematics, University of Oxford}
\affil[4]{Big Data Institute, University of Oxford}
\date{July 2020}

\begin{document}
\maketitle

\begin{abstract}
The most successful methods such as ReLU transfer functions, batch normalization, Xavier initialization, dropout, learning rate decay, or dynamic optimizers, have become standards in the field due, particularly, to their ability to increase the performance of Neural Networks (NNs) significantly and in almost all situations. Here we present a new method to calculate the gradients while training NNs, and show that it significantly improves final performance across architectures, data-sets, hyper-parameter values, training length, and model sizes, including when it is being combined with other common performance-improving methods (such as the ones mentioned above). Besides being effective in the wide array situations that we have tested, the increase in performance (e.g. F1) it provides is as high or higher than this one of all the other widespread performance-improving methods that we have compared against. We call our method Population Gradients (PG), and it consists on using a population of NNs to calculate a non-local estimation of the gradient, which is closer to the theoretical exact gradient (i.e. this one obtainable only with an infinitely big data-set) of the error function than the empirical gradient (i.e. this one obtained with the real finite data-set).
\end{abstract}


\section{Introduction}

Arguably, some of the methods that are having the deepest impact in AI are those that substantially improve performance (F1 or equivalent) in most or all situations \cite{reviewMethods}. Namely, methods that improve performance independently of the architecture used, the data-set, the length of training, the size of the model, or which other performance improving methods are being used in parallel. Methods such as ReLU transfer functions \cite{ReLU}, batch normalization \cite{BN_batchNormalisation}, Xavier initialization \cite{Xavier}, dropout \cite{dropout}, learning rate decay, or dynamic optimizers (e.g. Adam \cite{Adam_Adamax}, Adagrad \cite{Adagrad}, ...), are almost invariably applied, to the point that many of them have became the default in the most common libraries, such as PyTorch or Tensorflow, and now you actually need to explicitly instruct your NN library if you don't want your model using them.

Here, we present a novel method that improves performance in such a robust manner. The method consists on creating a population of noisy Neural Networks (NNs), with weights defined near the hyper-parameter position of the central NN that is being trained. The gradients of these NNs are then combined and back-propagated through the central NN, whose weights are then updated following this combined gradient. The hypothesis is that, by borrowing information on the gradients non-local to the central NN, a higher quality composite gradient is obtained, which avoids local minima and random fluctuations in the error function and its gradient. 

Based on the mechanism it exploits, we propose calling this method Population Gradients (PG). The method can be interpreted as an approximation to the use of higher order derivatives, which obtain information on the shape of the error function non-local to the current hyper-parameter position of the NN, although at prohibitive computational cost \cite{secondGradient}. For instance, second order derivatives or the Hessian matrix can indicate whether the local hyper-parameter position is a minimum, maximum or saddle point, but its computational cost prevents from using them in practice.

\section{Methods}

\subsection{Data-sets}

\begin{figure}[t]
\centering
\includegraphics[ trim = 2cm 2cm 0.5cm 2cm, clip, width = 8cm ]{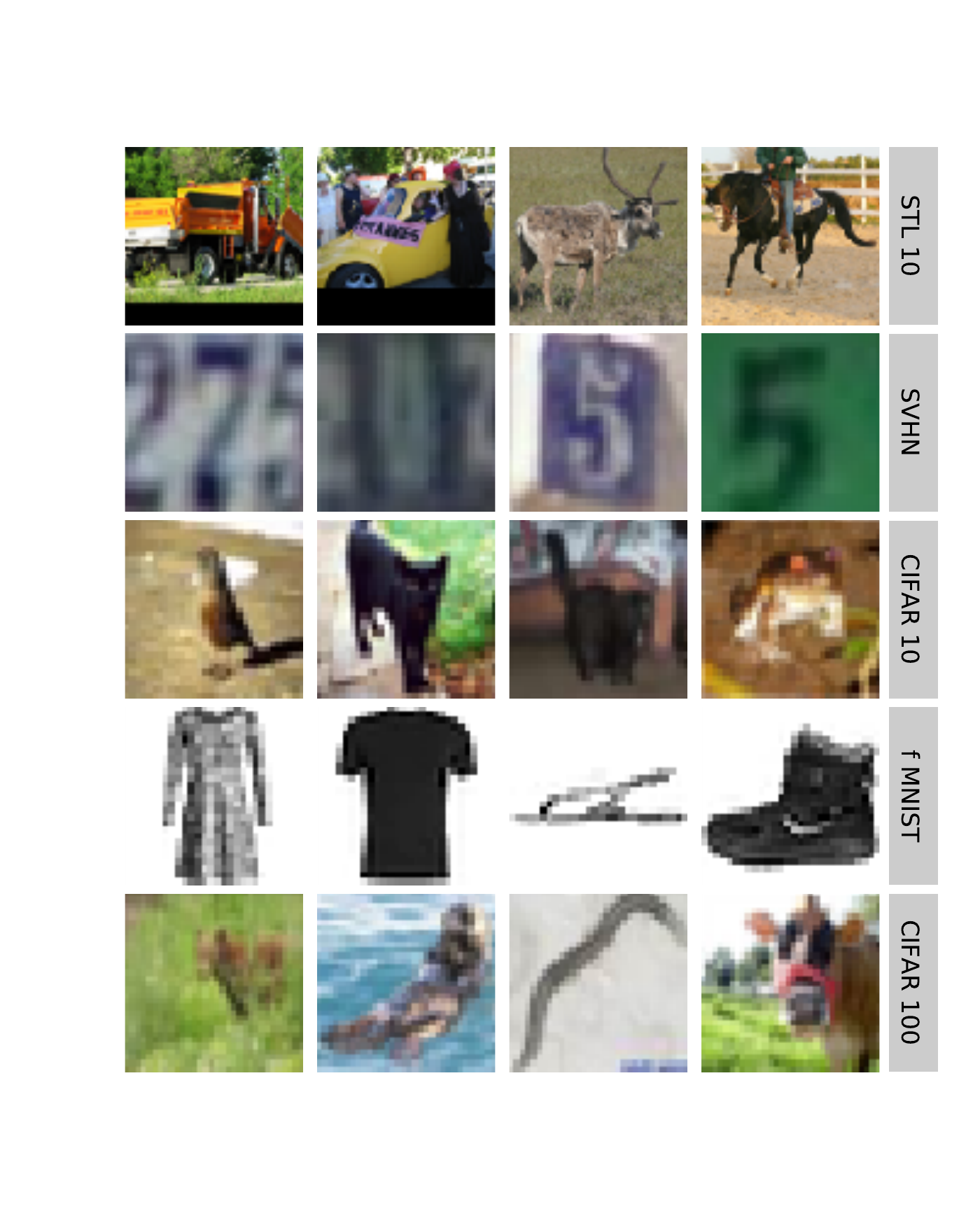}
\caption{ \label{fig_data} Data-sets. The figure shows 4 samples from each data-set, randomly selected from each training set. }
\end{figure}

If we were to conduct our experiments in a single data-set, the question would remain whether our results were generalizable to other data-sets. To answer this question, our experiments were conducted across 5 standard data-sets that comprehensively represented the space of object classification tasks. The data-sets were  CIFAR 10 \cite{CIFAR}, CIFAR 100 \cite{CIFAR}, fashion MNIST \cite{fMNIST}, STL 10 \cite{STL10}, and SVNH \cite{SVHN}. The characteristics of each data-set are detailed in Table \ref{table_data}, and examples of each given in Fig. \ref{fig_data}. Training and testing were always executed on the officially pre-extablished training and testing sets, respectively.

\begin{table}[t]
\centering
\begin{tabular}{ m{2.0cm} x{1.5} x{1.5} x{1.5} }
 &  \multicolumn{ 2 }{ c }{ \textbf{ samples - classes } } & \\
\textbf{data-set} & \textbf{training} & \textbf{testing} & \textbf{image size} \\
 \tableHline
 \textbf{ CIFAR 10 } & 50000 - 10 & 50000 - 10 & 32 x 32 x 3 \\
 \textbf{ CIFAR 100 } & 50000 - 100 & 50000 - 100 & 32 x 32 x 3 \\
 \textbf{ fashion MNIST } & 60000 - 10 & 60000 - 10 & 28 x 28 x 1 \\
 \textbf{ STL 10 } & 5000 - 10 & 8000 - 10 & 96 x 96 x 3 \\
 \textbf{ SVHN } & 73257 - 10 & 26032 - 10 & 32 x 32 x 3 \\
 \tableHline
\end{tabular}
\vspace*{ \tableMarginBottom }
\caption{\label{table_data} Data-sets. \upshape{ Characteristics of the data-sets used across experiments. Image sizes given in pixels and number of color channels. } }
\end{table}

\subsection{Baseline models}
\label{subsection:methods_architectures}


The question may also remain whether our results only hold for a particular NN architecture, or whether they generalize to other types of deep NN. To demonstrate generalizability across models, our experiments were conducted on 5 state of the art baseline architectures. These were DenseNet 161 \cite{DenseNet}, MobileNet v2 \cite{MobileNet}, ResNet 18 \cite{ResNet}, ResNet 50 \cite{ResNet}, and ShuffleNet v2 \cite{ShuffleNet}. All models were implemented, trained and tested in PyTorch. Training used a SGD optimizer with momentum 0.9, and weight decay $10^{-5}$. This momentum and weight decay almost always gave better performance than the PyTorch default values of 0.0.

\begin{table}[t]
\centering
\begin{tabular}{ m{2.5cm} x{3.5} }
 \textbf{ hyper-parameter } & \textbf{ tested values } \\
 \tableHline
 \textbf{ initial LR } & 0.002, 0.005, 0.01, 0.02, 0.05 \\
 \textbf{ batch size } & 64, 128, 256, 512 \\
 \textbf{ pads length } & 0, 1, 2, 4, 8 \\
 \textbf{ pads type } & zeros, border, reflection \\
 \textbf{ h-flip } & false, true \\
 \textbf{ norm mean } & 0.1, 0.2, 0.4, 0.8, None \\
 \textbf{ norm SD } & 0.1, 0.2, 0.4, 0.8, None \\
 \tableHline
\end{tabular}
\vspace*{ \tableMarginBottom }
\caption{\label{table_hyperparameters} Baseline hyper-parameters. \upshape{ The table lists all the values tested for each of the hyper-parameters of the baseline models. LR: learning rate; pads: padding, measured in pixels; h-flip: horizontal flip; SD: standard deviation. } }
\end{table}

\subsection{Tuning baseline models}
\label{subsection:methods_tuning}

\begin{figure}[t]
\centering
\hspace{ -0.25cm }
\subfloat[Calculating population gradients]{
    \includegraphics[ trim = 0.5cm 1.0cm 1.0cm 3.0cm, clip, height = 4cm]{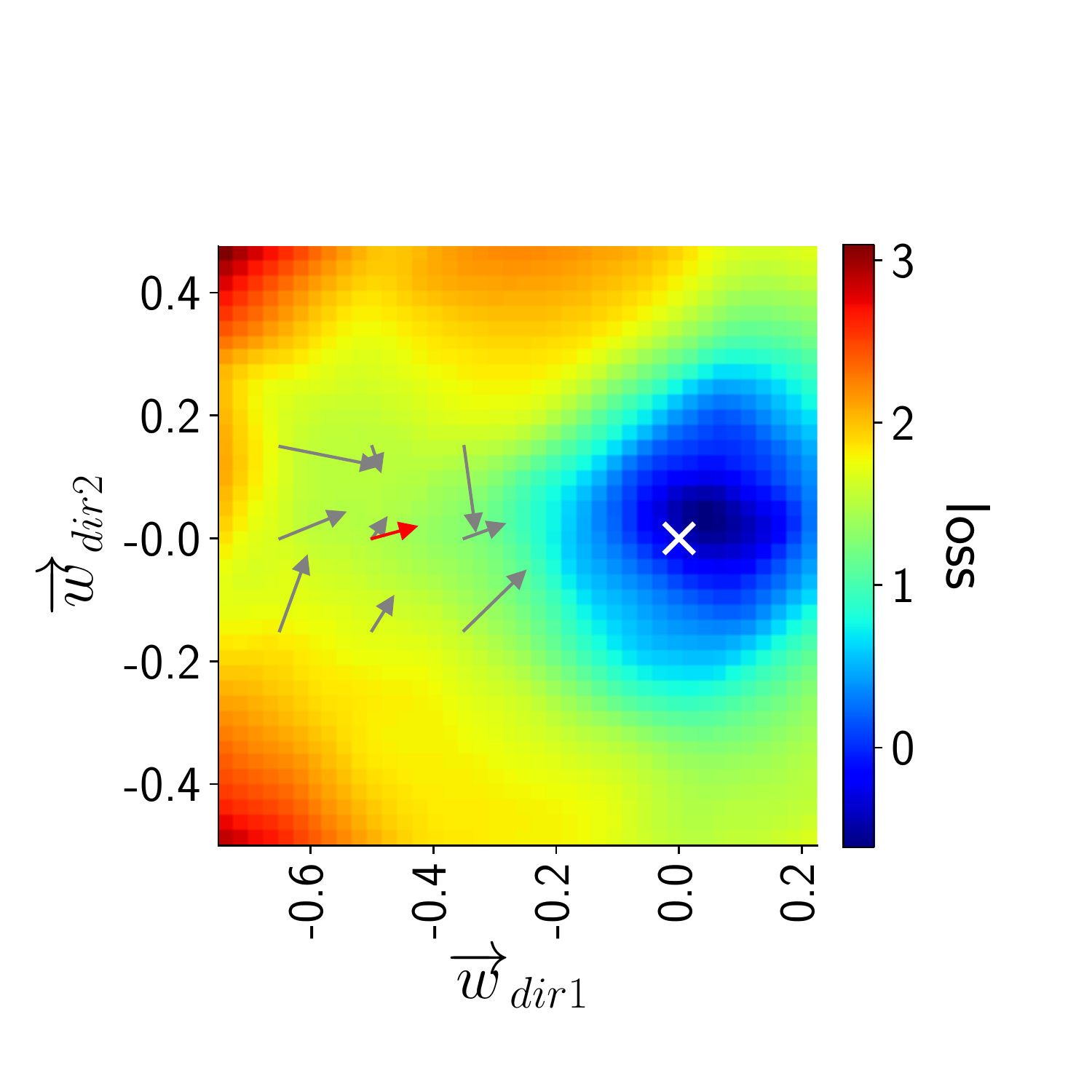}
    \label{fig_grad} }
\subfloat[Tuning baseline models]{
    \includegraphics[ trim = 0cm 0cm 0cm 0cm, clip, height = 4cm ]{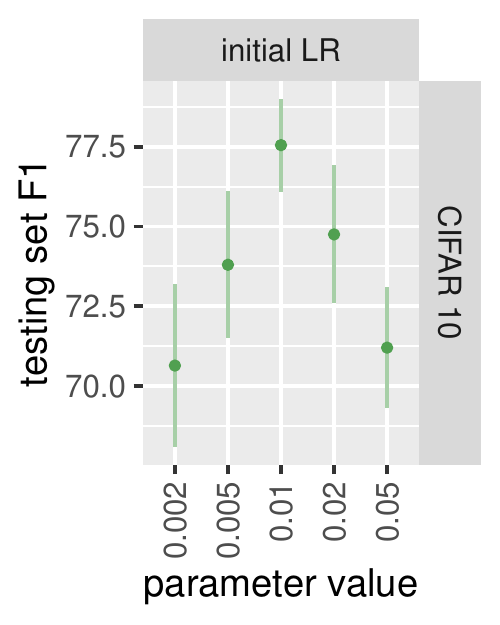}
    \label{fig_base} }
 \caption{ (a) Population Gradients. The figure depicts how population gradients are calculated in a real example. Background color shows the cross-entropy loss of ResNet 18 on a random batch of 32 samples of CIFAR 10. White cross locates the parameter values found after 100 epochs of training. Grey arrows show standard gradients near the point $( x = 0.5, y = 0.5 )$, while the red arrow show its average. For the sake of representing this in a figure, all these elements are projected in a 2D place cutting through the parameter space of the ResNet 18. x-axis is a random direction in the parameter space, while Y-axis is another random direction perpendicular to the first one. These random directions follow the same form of equation as the one used to calculate PG (equation \ref{eq:PG} ) - namely, $w_{trained} \cdot \big( 1 + x_{axis} \cdot w_{dir1} + y_{axis} \cdot w_{dir2} \big)$.
 (b). Tuning baseline models. The figure shows average performance (y-axis) of baseline architectures for different values of hyper-parameter "initial LR" (learning rate at the start of training). Averages (green dots) and standard deviations (green bars) are calculated across all evaluations of all baseline architectures when training on CIFAR 10, official training/testing split of samples used. }
\end{figure}

Both the architecture and the training process used have a number of hyper-parameters whose values impact the performance of the final NN (e.g. see Fig. \ref{fig_base}). If not investigated properly, two related questions would remain regarding how these values were chosen. The first question would be whether our results only hold true for a particular combination of hyper-parameter values, which we may have purposely chosen because PG performed best with such combination. The second question would be whether our results do outperform the baselines for the hyper-parameter values where these baselines perform best, or whether we have chosen values were such baselines perform poorly in order for PG to easily outperform the baselines. To address these two questions, we automatically searched the hyper-parameter space and selected the values with which the baseline model performed best. This process was repeated independently for each architecture $\times$ data-set, such that the values selected were those that performed best in that particular architecture $\times$ data-set. All subsequent experiments of that architecture $\times$ data-set, both those using PG and those not using PG, applied these same fine-tuned hyper-parameter values. Parameters not included in the search were always set to PyTorch defaults, unless stated otherwise. Training lasted for 100 mini-batch epochs, unless stated otherwise. Each epoch trained on the full official training data-set, which was divided into mini-batches without repetition of samples across them.

The search explored the following hyper-parameters: initial learning rate (LR), batch size, length of padding, padding type, horizontal flip, normalization mean and normalization standard deviation. Three of these parameters (3rd to 5th in the list) controlled how input data was augmented, while other two (latter 2 in the list) controlled how input images were normalized. Length of padding is the maximum number of pixels (and therefore padding) that input images were randomly shifted in the x and y directions. Pads type is the type of padding applied to the out-of-image pixels that emerge when images are randomly shifted. This can be 'zeros' (extra pixels are given the value 0), 'border' (pixels are given the colour of the closest inside-image pixel), and 'reflection' (the image is reflected around its rim to fill the extra pixels). H-flip indicates whether images were randomly flipped in the horizontal direction, with 50\% probability of being flipped or left as the original. Norm mean indicates the average value per color channel given to the input images, 'none' denoting when mean normalization was not applied. Norm SD indicates the standard deviation per color after normalization, 'none' denoting when SD was not controlled after normalization. When these latter two values equal 'none', the final effect is equivalent to not applying normalization at all. 

\begin{figure*}
\centering
\includegraphics[ trim = 0cm 0cm 0cm 0cm, clip, width = 17cm ]{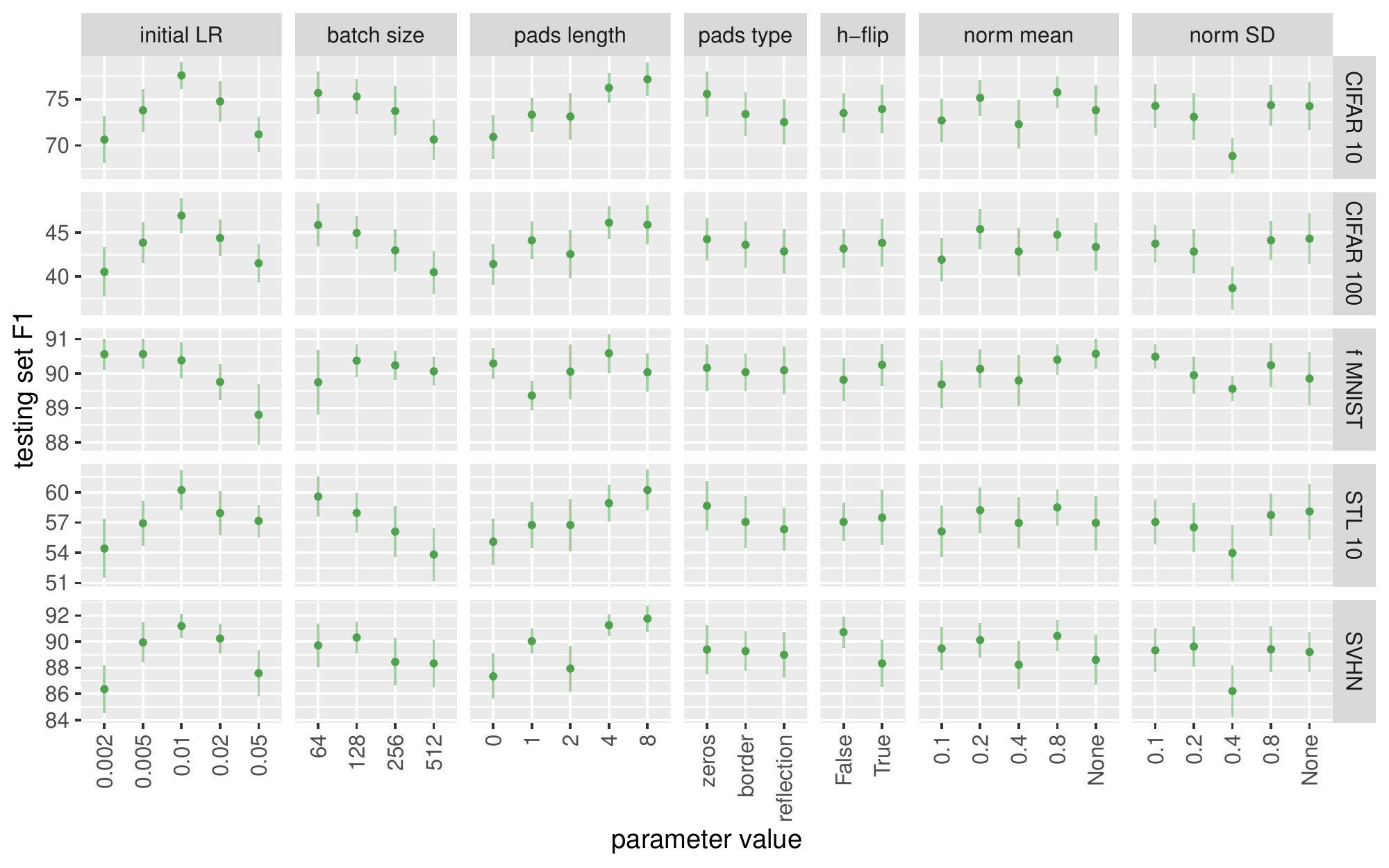}
\caption{ \label{fig_hyperF1} Tuning baseline models. The figure shows average performance (y-axis) of baseline architectures for different values of each explored hyper-parameter (columns of panels). Averages (green dots) and standard deviations (green bars) are calculated across all evaluations of all baseline architectures when training on a give data-set (rows of panels). Official training/testing split of samples used. }
\end{figure*}

The search explored a grid defined by a series of possible values for each of the hyper-parameters listed above. The possible values were those shown in table \ref{table_hyperparameters}. Parameters not in this list were given the default values defined in PyTorch, except for momentum and weight decay of SGD, which were set to values that achieved better performance than the PyTorch defaults. 1000 random combinations of hyper-parameter values within this search grid were evaluated, and their testing F1 score recorded after 100 iterations of mini-batch training. Samples were divided into training and testing sets as officially pre-established for each data-set. The average F1 was calculated independently for each combination of architecture $\times$ data-set, and used thereafter thorough the study. The hyper-parameter values that this search found to achieve highest test-set F1 per architecture $\times$ data-set are listed in table \ref{table_hyperparameters}, and the average test-set F1s obtained by each are shown in Fig \ref{fig_hyperF1}.

\subsection{Population gradients}
\label{subsection:methods_PG}

PG aims at increasing the quality of the gradients that are calculated with back-propagation. Here we introduce the concept of "gradient quality" as the ability of the "empirical gradient" (understood as the statistical average obtained from applying backpropagation to a finite number of samples) to approximate the "theoretical true gradient" (understood as the statistical mean obtained from applying backpropagation to a infinite number of samples). For instance, if our task were to classify pictures of cats and dogs, the empirical gradient would be the one obtained from a data-set of (e.g.) 1000 pictures, while the theoretical true gradient would be the one obtained from a theoretical data-set of infinity numbers of pictures. We hypothesize that, the closer an empirical gradient is to the true gradient, the better a NN will learn a task and generalize its performance to unseen samples. Although with this definition it is impossible to exactly calculate the quality of a gradient, it may be possible to estimate an approximation. The estimation would consist on calculating the average gradient with a subset of the available samples (e.g. 100 in the example above), and then compare this gradient to the one obtained with the whole data-set (e.g. 1000 samples). The comparison can be repeated multiple times (e.g. 10 if you use 100 samples each time in the example above) to also obtain the variance of this estimator.

A trivial avenue to improving the quality of a gradient consists on increasing the number of samples, but in practice this is often either not possible (e.g. in a medical data-set where you cannot "create" more patients), or it can only be exploited to a limited degree (e.g. once your company or grant runs out of funds to collect and annotate more pictures of cats and dogs). Rather than by increasing the number of samples, PG attempts at improving gradient quality by increasing the number of times that backpropagation is calculated. Arguably, simply calculating backpropagation multiple times on the same samples (i.e. batch) won't improve the quality of gradients, as you are not providing additional information. Rather, an additional but informative source of variability needs to be incorporated into backpropagation. When you increase the number of samples, this additional but informative source of variability are the new samples themselves. PG brings this source of variability from slightly modifying the values of the weights of the neural network each time backpropagation is applied. Algorithmically, for each sample we calculate backpropagation 's' times, incorporating a noise term of standard deviation 'w' into the weights of the NN each time. The parameter 's' is called 'population size', and 'w' 'population range'. Any trained NN architecture is expected to have weights of very different magnitudes. To control for this large magnitude disparity, noise is applied through multiplication rather than through summation. Mathematically, noise is applied to each weight as follows:
\begin{equation}
w_{final} = w_{original} \cdot \big( 1 + \mathcal{N}( 0, r ) \big)
\label{eq:PG}
\end{equation}
where $\mathcal{N}( 0, r )$ is a random number drawn from a normal distribution of standard deviation 'r' (population range)

\subsection{Evaluating population gradients}
\label{subsection:methods_evaluatingPG}

In another interpretation, we can understand PG as a method that creates a population of 's' noisy NN around the original one. These noisy NNs compute the neighbouring gradient non-local to the original one. Combining these non-local gradients brings information on the shape of the gradient around the original NN. This should reduce random fluctuations of the gradient, and average away local minima, potentially improving the quality of the gradient. The question however remains whether this process actually improves the quality of the gradient and the final performance of a NN after training. Proving this is the case across most situations is the objective of this study.

As defined above, PG has two meta-parameters (population size 's', and population range 'r') that regulate its behaviour, and their value is expected to impact the final performance of the NN model. To document their effect, we evaluated the performance of the NN for each combination of a series of possible values per meta-parameter. These values were { 5, 10 } for population size, and { 0.05, 0.1, 0.2, 0.4 } for population range. On each evaluation the average testing F1 score was recorded after 100 iterations of mini-batch training. Besides these two meta-parameters, all evaluations used the hyper-parameter values calculated in section \ref{subsection:methods_tuning}. listed in table \ref{table:tuning}. It is important to note that these hyper-parameter values were optimized for the baseline models, not for the models while using PG. 

\begin{figure*}[h]
\centering
\includegraphics[ trim = 0cm 0cm 0cm 0cm, clip, width = 17cm ]{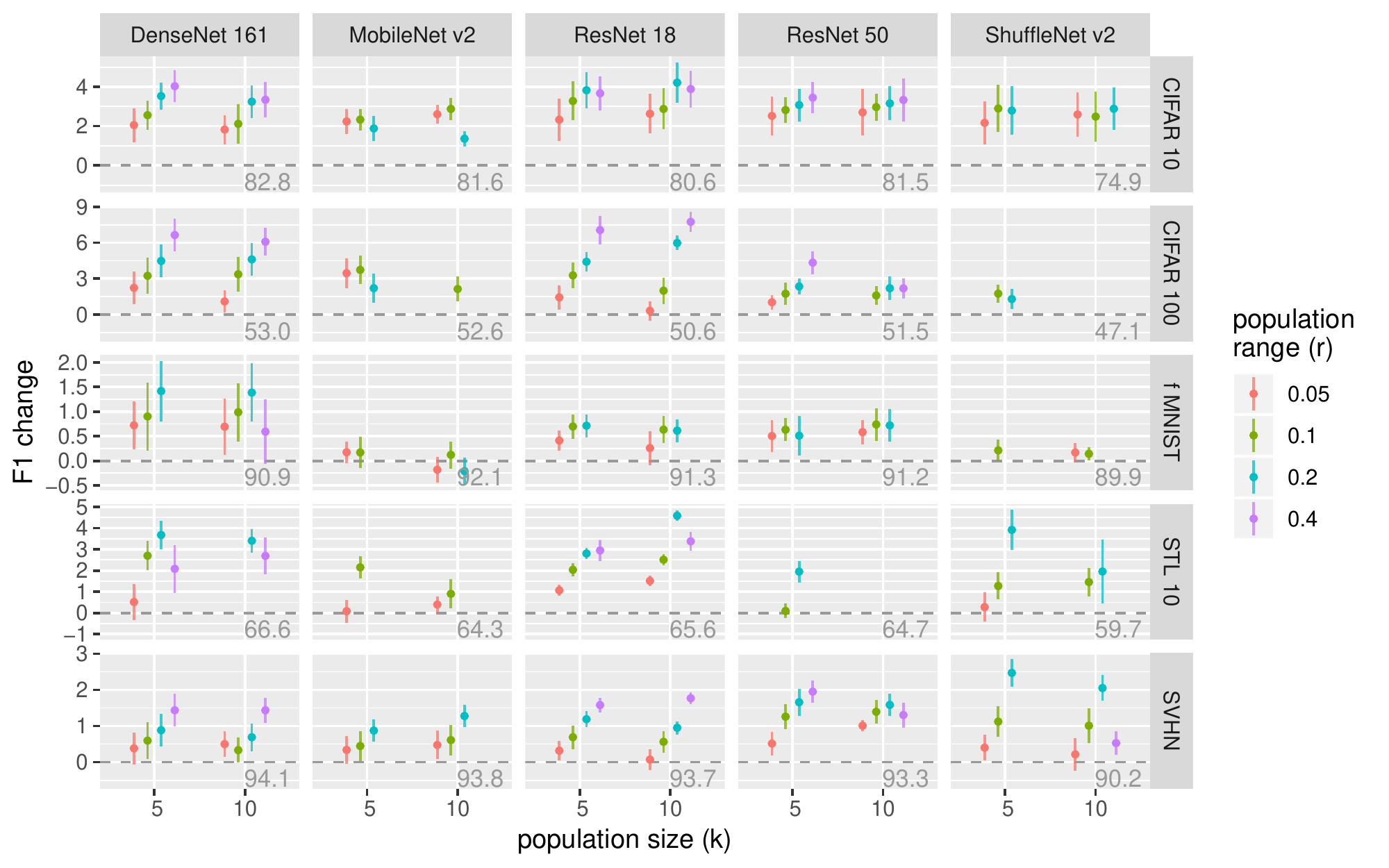}
\caption{ \label{fig_pop2} Evaluating population gradients. The figure shows performance improvement (y-axis, testing set F1) of baseline architectures when these use PG. This is shown separately for each architecture (columns of panels), data-set (rows of panels), population size ('k', x-axis), and population range ('r', color). In each case, average improvement (dots) and standard deviations (bars) are calculated across the last 10 epochs of training. The lower right corner of each panel shows the performance of the baseline NN for that architecture $\times$ data-set (grey numbers within panel). Official training/testing splits of samples used. [YYY] Update using the new code that makes the dashed lines equidistant. }
\end{figure*}

\subsection{Comparing to other methods}
\label{subsection:methods_comparing}

The experiments listed above measure how the performance of fine-tuned baseline models (i.e. base line models using the hyper-parameter values of section \ref{subsection:methods_tuning}) increases when PG is applied. The question remains regarding how this increase compares to other methods that also aim at improving performance across architectures and data-sets. To demonstrate this comparison, we selected some of the most prominent methods: LR decay, dropout, L1 regularisation, L2, Nesterov's optimization, RMS-Prob, Adam-W, Adam, AMS-Grad, and Adamax.

Each of these additional methods also have some meta-parameters that impact the final performance of the NN model. Even if we compare PG to all these other popular methods, the question may remain whether we were using meta-parameter values that performed poorly for these methods, such that PG easily outperformed them. To demonstrate this was not the case, we evaluated each method for each combination of possible values per method's meta-parameter. Further, in all cases we tested more values than those explored for PG. The evaluated values are those listed in Table \ref{table_comparing}

\begin{center}
\begin{table}
\centering
\begin{tabular}{ m{ 1.5 cm } x{ 1.75 } x{ 3.5 } }
\textbf{ method } & \textbf{ method's parameters } & \textbf{ tested values }  \\
 \tableHline
 \multirow{2}{*}{ \textbf{ PG } } & pop. size & { 5, 10 } \\
      & pop. range & { 0.05, 0.1, 0.2, 0.4 } \\
 \tableHline
 \multirow{2}{*}{ \textbf{ LR decay } } & number of eras & 2, 5, 10, 20 \\
      & era decay & 0.9, 0.7, 0.5, 0.2 \\
 \tableHline
 \multirow{2}{*}{ \textbf{ dropout } } & 1st layer & 0.0, 0.1, 0.2, 0.4 \\
      & last layer & 0.0, 0.1, 0.2, 0.4 \\
 \tableHline
  \multirow{2}{*}{ \textbf{ L1 reg. } } & 1st layer & $10^{-5}$, $10^{-4}$, 0.001, 0.01, 0.1 \\
      & last layer & $10^{-5}$, $10^{-4}$, 0.001, 0.01, 0.1 \\
 \tableHline
  \multirow{2}{*}{ \textbf{ L2 reg. } } & 1st layer & $10^{-5}$, $10^{-4}$, 0.001, 0.01, 0.1 \\
      & last layer & $10^{-5}$, $10^{-4}$, 0.001, 0.01, 0.1 \\
 \tableHline
 \multirow{2}{*}{ \textbf{ Nesterov's } } & momentum & 0.1, 0.5, 0.9, 0.99 \\
      & dampening & 0.991, 0.995, 0.999, 0.9999 \\
 \tableHline
 \multirow{2}{*}{ \textbf{ RMS-Prob } } & momentum & 0.1, 0.5, 0.9, 0.99 \\
      & alpha & 0.991, 0.995, 0.999, 0.9999 \\
 \tableHline
 \multirow{2}{*}{ \textbf{ Adam-W } } & beta 1 & 0.1, 0.5, 0.9, 0.99 \\
      & beta 2 & 0.991, 0.995, 0.999, 0.9999 \\
 \tableHline
 \multirow{2}{*}{ \textbf{ Adam } } & beta 1 & 0.1, 0.5, 0.9, 0.99 \\
      & beta 2 & 0.991, 0.995, 0.999, 0.9999 \\
 \tableHline
 \multirow{2}{*}{ \textbf{ AMS-Grad } } & beta 1 & 0.1, 0.5, 0.9, 0.99 \\
      & beta 2 & 0.991, 0.995, 0.999, 0.9999 \\
 \tableHline
 \multirow{2}{*}{ \textbf{ Adamax } } & beta 1 & 0.1, 0.5, 0.9, 0.99 \\
      & beta 2 & 0.991, 0.995, 0.999, 0.9999 \\
 \tableHline
\end{tabular}
\vspace*{ \tableMarginBottom }
\caption{\label{table_comparing} Method's meta-parameters. \upshape{ Column "method's parameters" lists the names of the parameters that were fine tuned for each method, while column "evaluated values" lists the values that were explored for each of these parameters. pop.: population; reg.: regularisation. } }

\end{table}
\end{center}

What meta-parameters, if any, are incorporated into each of these prominent methods may vary from study to study. Here we used in each method the 2 parameters that, we believe, best explored its range of possible behaviours:

\begin{itemize}
\item For LR decay, we divided training into a number of eras of equal length. LR remained constant within each era, and was set to the standard LR (i.e. the fine-tuned LR selected in section \ref{subsection:methods_tuning}, table \ref{table:tuning}) for the first era. At the end of each era, LR was multiplied by a decay factor, with this factor being called 'era-decay'.
\item For dropout, we added a drop-out layer before each activation function of the original baseline architecture. We gave to the first dropout layer a fixed dropout probability, and another fixed probability to the last layer. The dropout probability of each intermediate layer was then established by a linear interpolation between the probabilities of the first and last ones. Namely, if $y(x)$ is the probability of each layer $x$, the interpolation consisted on calculating the values of $b$ and $x_0$ in the linear equation $y(x) = bx + x_0$ such that $y(1)$ matched the probability of the first layer, and $y(x_{last})$ the probability of the last layer (with $x_{last}$ being the total number of layers of the architecture). Each intermediate layer was then given the probability $\{y(2), y(3), ..., y(x_{last}-1)\}$. As an example, in a NN of 5 layers with $y(1) = 0.6$ and $y(x_{last}) = 0.2$, $y(x)$ would be $\{0.6, 0.5, 0.4, 0.3, 0.2\}$.
\item For L1 and L2 regularisation, we calculated the average activation of each layer of the baselines architecture. The average was calculated independently in each layer to avoid bias towards the layers with the highest number of neurons. These averages were multiplied by a layer-wise regularisation factor, then averaged again, and the result added to the final loss of the NN. As with dropout, we defined a 1st layer and a last layer regularisation factor, with all intermediate ones being interpolated between these.
\item For each advanced optimizer, we used the two standard parameters that are built into their original PyTorch implementation. These optimizers were Nesterov's, RMS-Prob, Adam-W, Adam, AMS-Grad, and Adamax. See table \ref{table_comparing} for the list of meta-parameters. When applied, these optimizers replaced the default SGD, which was used otherwise.
\end{itemize}

\subsection{Measuring effect of combining with other methods}
\label{subsection:methods_combining}

Besides fine tuning their hyper-parameters, baseline architectures such as the ones tested in this study are often combined with multiple performance improving methods in order to enhance their accuracy. The question therefore remains whether PG can also be combined with other methods in this manner, or whether it only works when used in isolation.

To investigate whether PG further increases the performance when combined with other prominent performance-increasing methods (rather than both methods cancelling the positive effects of each other), we evaluated the performance of PG when applied to NN that already used each of the alternative methods of section \ref{results_comparing}. To save computation time, we tested such combination only in one of the architecture $\times$ data-sets (ResNet 18 $\times$ CIFAR10), assuming results are generalizable to other architectures and data-sets. We believe this is a safe assumption given that PG increases performance across all 25 tested architecture $\times$ data-sets (see section \ref{subsection:results_baselines}). In all cases, we used the tuned hyper-parameter values that performed best for the original baseline NN model, as calculated in section \ref{subsection:methods_tuning}, listed in table \ref{table_metaparameters}. In order to save time, we did not fine-tuned the meta-parameters of PG, and rather used those that performed best for ResNet 18 $\times$ CIFAR10 in section \ref{subsection:methods_PG}. We however fine-tuned again the meta-parameters of each of the other alternative method, using the same grid search as in section \ref{subsection:methods_comparing}. Their meta-parameters were fined tuned again when each method was combined with PG.

\subsection{Measuring effect of long training}
\label{subsection:methods_long}

Some recent publications are demonstrating that, contrary to classical belief, very long training (i.e. an unusually high number of epochs) actually improves the performance of NNs \cite{doubleDescend}. Up until now, the long held believe in machine learning (ML) had been that, in any ML model, training increases only up to an optimal point, after which further training damages performance due to over-fitting \cite{bishop, probabilisticMachineLearning}. A long collection of theoretical work further supported this belief, arguing that up to the optimal point the ML model learned to average the variance of the data, and that after that optimal point the ML model started to overuse its inherent bias. Empirical observations also seemed to always confirm this belief. Publications often showed this effect in NN as well. However and surprisingly, at least in the case of NNs, if training is continued for an unusually long period, performance starts rising again, and often even surpasses the previous optimal point after enough training. This so-called 'double descend' is however only observed in the larger type of architectures and data-sets, which may explain why it was not observed in traditional ML models, given that their limited expressive power is more akin small NNs. These recent result suggest that protracting training for very long periods may become a new default method towards increasing model performance.

In light of these recent results, the question arises whether PG also improves performance after very long training. The ability to further increase performance when combined with very long training would further increase the worth of PG as a performance increasing method, as exploiting the double gradient descent may soon become common practise. To test this possibility, we evaluated PG in ResNet 18 $\times$ CIFAR 10. We used the hyper-parameter values of section \ref{subsection:methods_tuning} (i.e. those that performed best for the original baseline model without PG), and the meta-parameters values of section \ref{subsection:comparing} (i.e. those that performed best for PG).

\section{Results}

\subsection{Evaluating population gradients}
\label{subsection:results_baselines}

As described in methods, we first evaluated the performance that 5 different baseline architectures (DenseNet 161 \cite{DenseNet}, MobileNet v2 \cite{MobileNet}, ResNet 18 \cite{ResNet}, ResNet 50 \cite{ResNet}, and ShuffleNet v2 \cite{ShuffleNet}, see section \ref{subsection:methods_architectures}) achieved in 5 different object classification data-sets (CIFAR 10 \cite{CIFAR}, CIFAR 100 \cite{CIFAR}, fashion MNIST \cite{fMNIST}, STL 10 \cite{STL10}, and SVNH \cite{SVHN}, see Table \ref{table_data}, and section \ref{table_data}). The baseline models were fine-tuned via random search (1000 evaluations) on the space defined by 7 hyper-parameters (see table \ref{table_hyperparameters}, or section \ref{subsection:methods_tuning}). We then evaluated the performance that each architecture $\times$ data-set achieved when PG was applied (see section \ref{subsection:methods_evaluatingPG}), while using the same fine-tuned hyper-parameters as used by the baseline model (i.e. we didn't fine-tuned the hyper-parameters again to find the values that best performed when PG was applied). The evaluation with PG was repeated for different values of 'population size' (parameter 's'), and 'population range' ('r'), the two internal parameters of PG that control its behaviour. We will refer to 's' and 'r' as meta-parameters to distinguish them from the 7 hyper-parameters of the baseline model (see section \ref{subsection:methods_tuning} ).

Fig. \ref{fig_pop2} shows the increase in performance that PG obtained in each architecture $\times$ data-set for all tested meta-parameters. The average maximum increase per architecture was 6.6 $\pm$ 2.7 (DenseNet 161, mean $\pm$ SD), 3.7 $\pm$ 2.4 (MobileNet v2), 7.7 $\pm$ 1.7 (ResNet 18), 4.3 $\pm$ 1.9 (ResNet 50), and 3.9 $\pm$ 1.9 (ShuffleNet v2). The overall average of averages was therefore 5.3 $\pm$ 1.6. The average maximum per data-set was  4.2 $\pm$ 2.0 (CIFAR 10), 7.7 $\pm$ 1.7 (CIFAR 100), 1.4 $\pm$ 1.2 (fashion MNIST), 4.6 $\pm$ 0.5 (STL 10), and  2.5 $\pm$ 0.8  (SVNH). The overall average of averages was 4.1 $\pm$ 2.2. The internal parameter values that most commonly achieved the best performance were 5 for 'population size' (best 19 times out of 25 architecture $\times$ data-set), and 0.2 for 'population range' (11 out of 25), closely followed by 0.4 (10 out of 25). 

\begin{figure}[t]
\centering
\includegraphics[ trim = 0cm 0cm 0cm 0cm, clip, width = 6cm ]{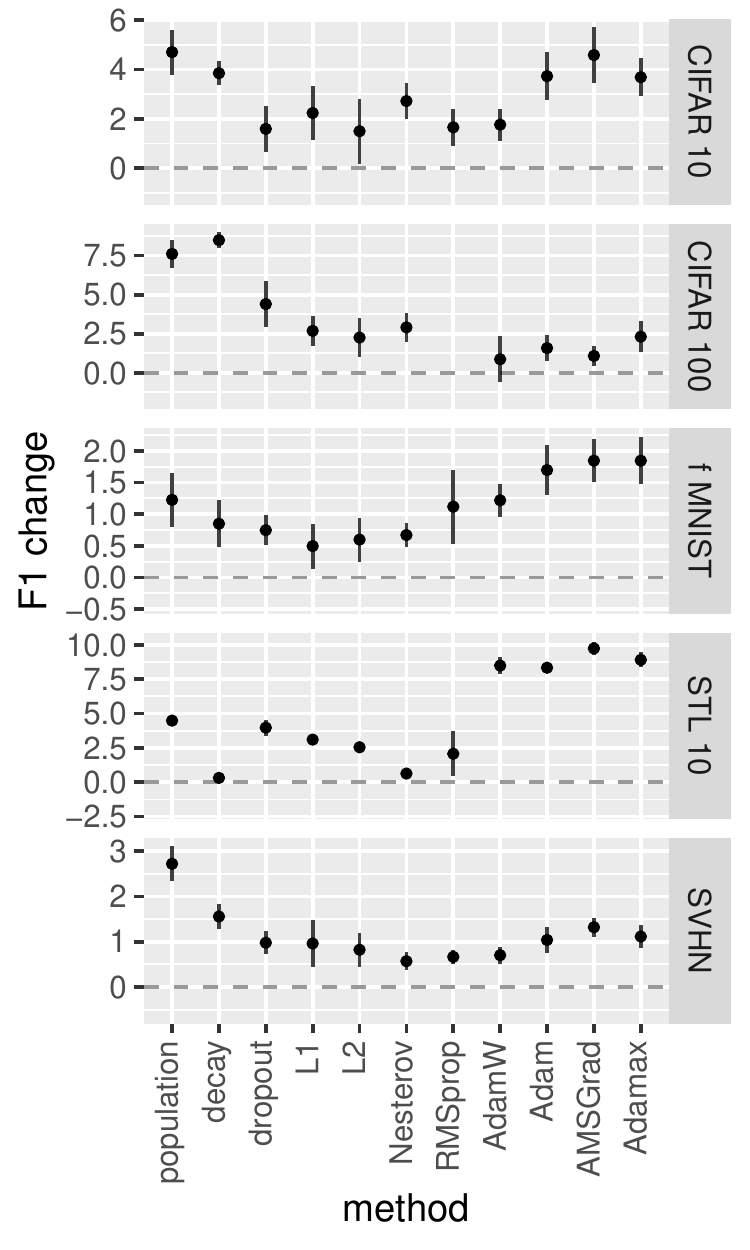}
\caption{ \label{fig_comp} Comparing to other methods. The figure shows maximum performance improvement (y-axis, testing set F1) per data-set when using each method (x-axis). In each case, averages (dots) and standard deviations (bars) are calculated across the last 10 epochs of training of the best model evaluation. The best model evaluation is the one that produced the best performance improvement (measured as the average of the last 10 epochs of training) across architectures and method's meta-parameters (method's parameters are listed in table \ref{table_comparing}). Official training/testing splits of samples used. }
\end{figure}

\subsection{Comparing to other methods}
\label{subsection:results_comparing}

After evaluating the increase in performance of PG across architectures and data-sets, we compared it to the increase achieved by other 10 major enhancing methods - LR decay, dropout, L1 regularisations, L2, Nesterov's optimization, RMS-Prob, Adam-W, Adam, AMS-Grad, and Adamax. The two most common meta-parameters of each of the alternative method and of PG were optimized via systematic grid search (see section \ref{subsection:methods_comparing}, evaluating the values listed in table \ref{table_metaparameters}. Each grid search tested either 8 different meta-parameter combinations (in the case of PG), or 16 (in the case of LR decay, dropout, and each of the advanced optimizers), or 25 (in the case of L1 and L2, in an attempt to improve the very low performance increase obtained with these methods).

Fig. \ref{fig_comp} and table \ref{table_comparing} show the maximum increase in performance that PG obtained in each data-set (across architectures), compared to the maximum increase achieved by each alternative method. This maximum increase was in average  4.8 $\pm$ 3.4 (F1 $\pm$ SD in the last 10 epochs of training, see figure \ref{fig_comp}) for PG, while for the other methods it was 3.0 $\pm$ 3.0 (LR decay), 2.3 $\pm$ 1.5 (dropout), 1.8 $\pm$ 1.1 (L1), 1.5 $\pm$ 0.8 (L2), 1.5 $\pm$ 1.1 (Nesterov's), 0.3 $\pm$ 2.2 (RMS-Prob), 2.6 $\pm$ 3.0 (Adam-W), 3.3 $\pm$ 2.7 (Adam), 3.6 $\pm$ 3.3 (AMS-Grad), and 3.7 $\pm$ 2.8 (Adamax). Therefore, the average increase of PG was larger than the average increase of all other methods, while the maximum increase across architectures per data-set was among the highest of all methods. Among all 11 methods, the rank of PG as a enhancing method across architectures was therefore 1st (CIFAR 10), 2nd (CIFAR 1000), 4th (fMINST), 5th (STL 10), and 1st (SVHN)

\subsection{Measuring the effect of combining with other methods}
\label{subsection:methods_combining}

After comparing with other methods, we measured the effect of combining PG with each of these alternatives. Again, the two meta-parameters of PG were not optimized further, and rather set to those that obtained the highest performance in section \ref{subsection:methods_evaluatingPG}. Meanwhile, the meta-parameters of each of the alternative methods were optimized again to each new experiment, using the same random grid-search approach as in section \ref{subsection:results_comparing} (i.e. 1000 tests randomly selected within the grid defined in table \ref{table_metaparameters}). To save computation time, these experiments were executed only on ResNet 18 with CIFAR 10. Given that optimizers are mutually exclussive (i.e. you can only use one of them to increase model performance), we excuted this experiment on the one that performed best in section \ref{subsection:results_comparing} - which was Adamax.

\begin{figure}[t]
\centering
\includegraphics[ trim = 0cm 0cm 0cm 0cm, clip, height = 5.5cm ]{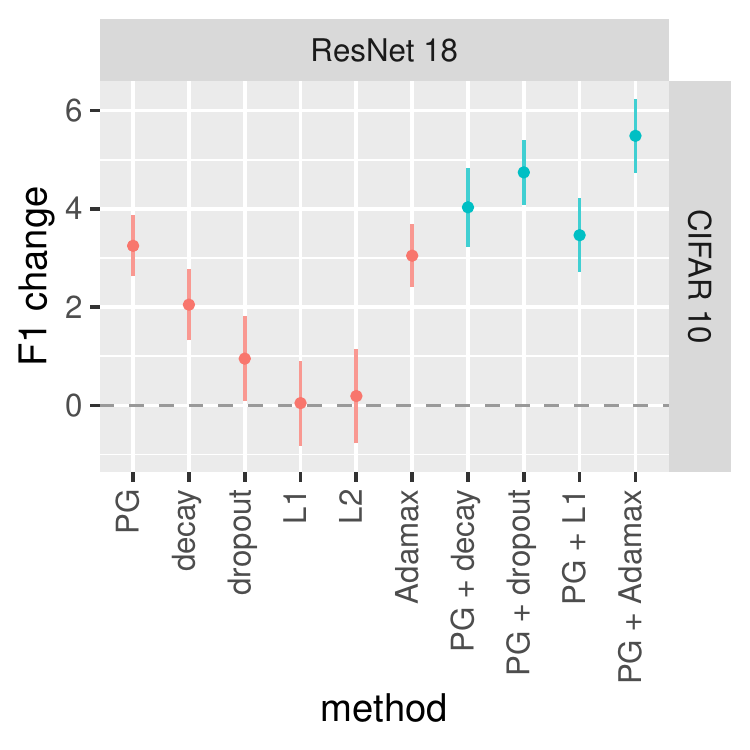} 
\caption{ \label{fig_comb} Combining with other methods. The figure shows maximum performance improvement (y-axis, testing set F1) of baseline ResNet18 when it uses each method in isolation or combined with PG (x-axis). Averages (dots) and standard deviations (bars) are calculated across the last 10 epochs of training of the best model evaluation. The best model evaluation is the one that produced the best performance improvement (measured as the average of the last 10 epochs of training) method's meta-parameters (method's parameters are listed in table \ref{table_comparing}). The meta-parameters of PG were held constant across evaluations. Official training/testing splits of samples used }
\end{figure}

In all cases, PG further increased the performance of each of the alternative methods (see Fig. \ref{fig_comb}). The final performance of the combination was also always higher than the performance of PG alone. When combined with PG, the increase with respect to baseline was 4.0 $\pm$ 1.6 (for LR decay), 4.7 $\pm$ 1.3 (for dropout), 0.0 $\pm$ 1.0 (for L1), 0.2 $\pm$ 0.9 (for L2), and 5.5 $\pm$ 1.5 (for Adamax). 

\begin{figure*}[h]
\centering
\includegraphics[ trim = 0cm 0cm 0cm 0cm, clip, width = 17cm ]{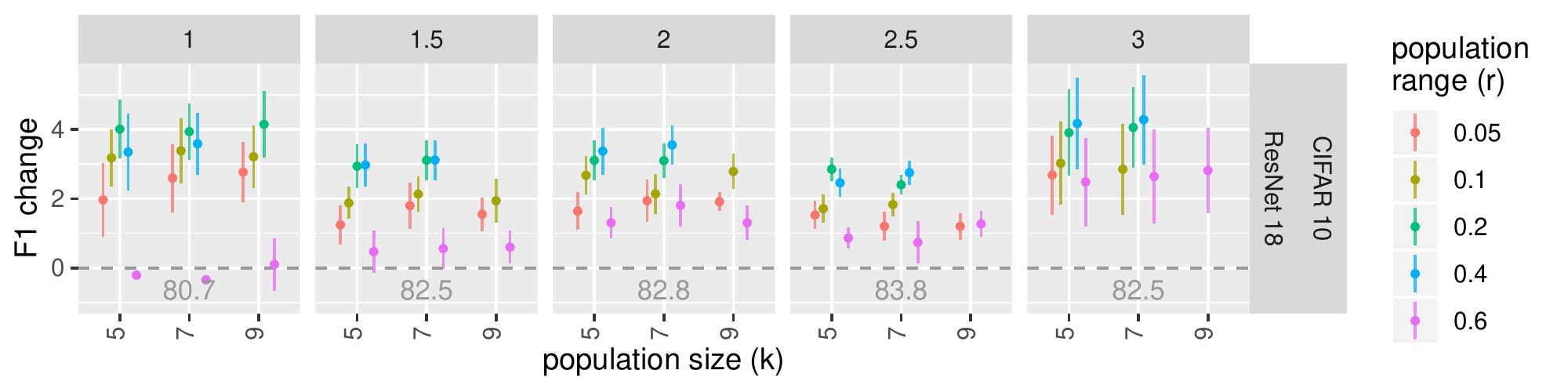} 
\caption{ \label{fig_big} Effect of large architectures. The figure shows the performance improvement (y-axis, testing set F1) of PG when increasing the size (number of channels per layer) of the baseline architecture by a multiplicative factor (columns of panels). Averages (dots) and standard deviations (bars) are calculated in non intersecting bins of 25 epochs. Official training/testing splits of samples used. The lower right corner of each panel shows the performance of the baseline NN for that multiplicative factor (grey numbers within panel). }
\end{figure*}

\begin{figure}[h]
\centering
\includegraphics[ trim = 0cm 0cm 0cm 0cm, clip, height = 5.5cm ]{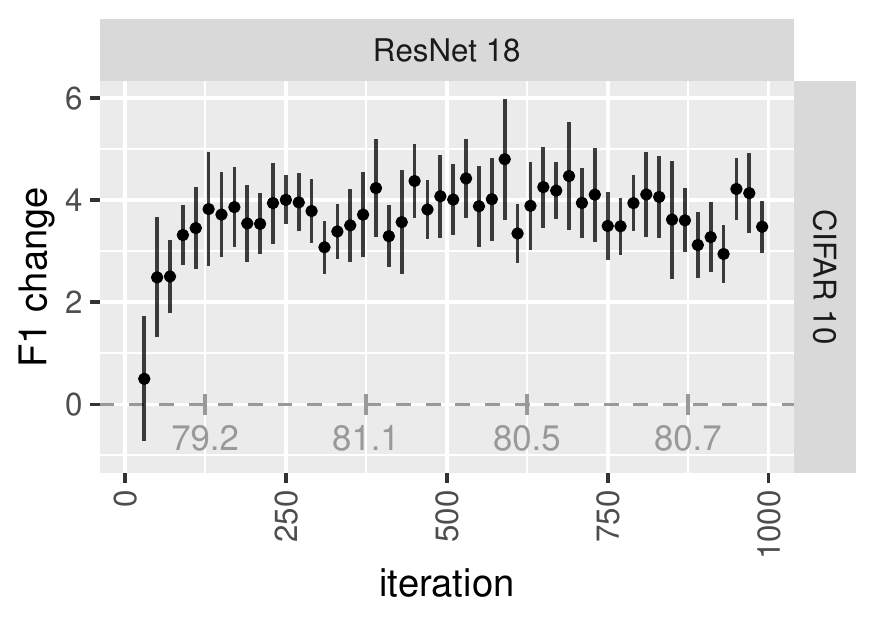} 
\caption{ \label{fig_long} Effect of long training. The figure shows the performance improvement (y-axis, testing set F1) of PG when training is continued beyond 100 iterations (x-axis). Averages (dots) and standard deviations (bars) are calculated in non intersecting bins of 25 epochs. Official training/testing splits of samples used. The lower edge of the panel shows the performance of the baseline NN averaged in bins of 250 epochs (grey numbers within panel)}
\end{figure}

\subsection{Measuring effect of long training}
\label{subsection:results_long}

Recent studies show that, rather than unbounded over-fitting, often there is a second increase in performance if NNs are trained for very long periods \cite{doubleDescend}. This makes very long training an attractive method to further increase performance, and PG would be all the more attractive if it demonstrate to be effective also in this scenario. To test this feature, we also measured the effect that very long training had on PG by using 10 times more epochs than in previous sections - namely 1000 epochs rather than 100. To acomodate for the long computation time that this required, we only tested on ResNet $\times$ CIFAR 10. Again, the two meta-parameters of PG were not optimized further, and rather set to those values that obtained the highest performance in section \ref{subsection:methods_evaluatingPG} for that architecture $\times$ data-set. As in previous sections, all experiments used the hyper-parameter values calculated in section \ref{subsection:methods_tuning}, which were optimal for the baseline architectures when PG was not active.

The following numbers are reported after averaging F1 values in non-intersecting windows of 20 epochs (i.e. 0-19, 20-39, 40-59... 980-999, totalling 50 bins). The baseline NN only increased performance up to 81.8 $\pm$ 1.0 after ~100 epochs (see figure \ref{fig_long}). This maximum occurred at the bin of epochs 300-319, and did not appear to increase further in later epochs. However, PG continued increasing F1, holded at an apparently near constant value of up to 4.8 $\pm$ 2.4 above the baseline. This maximum occurred at bin of epochs 580-599, and should be added to the 80.9 $\pm$ 2.4 that the baseline obtained in this same bin (namely 85.7 F1). PG achieved higher performance that the baseline in all 50 epoch bins.

\subsection{Measuring effect of large architectures}
\label{subsection:results_big}

The same study that demonstrated the benefit of training for very long periods \cite{doubleDescend}, also showed that very large architectures also often increase performance rather than always producing over-fitting. Again, this makes increasing architecture size an attractive method to further improve performance, and PG would be all the more attractive if it were effective also in this scenario. To test this feature, we also measured the effect of increasing architecture size up to 3.5 fold of the original size, which was the maximum memory capacity of each of our RTX 2080 TI GPUs. To accommodate for the long computation time, we also only tested on ResNet 18 $\times$ CIFAR 10. In an attempt to make the effect of the increase in size independent of the specific type of architecture, we expanded the architectures by multiplying the number of channels in each convolution layer, and of neurons in each non convolution layer. To more widely explore the behaviour of PG, we evaluated it in all combinations of population size (k) being 5, 7, 9, and population range (r) being 0.05, 0.1, 0.2, 0.4 and 0.6.  As in previous sections, all experiments used the hyper-parameter values calculated in section \ref{subsection:methods_tuning}, which were optimal for the baseline architectures when PG was not active. 

The baseline NN only increased performance up to 83.8 $\pm$ 0.6 (see Fig. \ref{fig_big}), which occurred at size $\times 2.5$. PG increasing F1 over baseline in all larger networks, up to 4.3 $\pm$ 2.6 at size $\times 3.0$. As before, this maximum should be added to the 82.5 $\pm$ 2.5 performance that the baseline obtained at this NN size (i.e. 86.8 F1), in order to account for the benefit of simultaneously enlarging the model and using PG. Interesting, rather than the maximum improvement occurring most often for $k=5$ and $r=0.2$, as it was the case in the original sized network (i.e. section \ref{subsection:results_baselines}), the maximum ensued often at $k=9$ (all sizes above $\times 1.0$) and $r=0.4$ (3 out of 4 sizes above $\times 1.0$), specially in the largest networks.

\section{Discussion}

We believe this study confidently establish PG as a robust method to improve performance in most situations. Namely, our results show that PG improves performance independently of architecture, data-set, training time, and model size. It also improves performance further when combined with other popular methods, such as LR decay, dropout, L1 regularisation, L2, and modern complex optimizers. The increase in performance is as high or higher than the increase achieved by these very popular alternative methods. If these alternative methods have become so universally used (to the point that they are now often the default in NN libraries), we believe PG will also become a comparably popular default addition to NNs.

We anticipate and encourage further work to further study the capabilities and limits of PG. For instance, it would be exciting to test how the method performs when: (1) it is used in other fields (e.g. NLP, generative models, reinforcement learning...); (2) its noise distribution is improved (e.g. noise with L1 regularisation - namely noise modifying a limited proportion of weights); (3) its behaviour refined (e.g. different parameters per layer); (4) its parameters better fine-tuned (e.g. fine-tuning simultaneously baseline hyper-parameters, PG meta-parameters and other methods' meta-parameters); (5) it is used in the most challenging tasks (e.g. using as baseline architectures the best performing state of the arts of each data-set \cite{papersWithCode}).

In summary, we believe PG is a notable contribution to the field, and encourage future work to further study the capabilities and limits of PG.

\appendices

\beginsupplement

\begin{table}
\centering
\begin{tabular}{ m{1.5cm} x{1.5} x{2.3} }
     & & \textbf{ max improvement } \\
    \textbf{ method } & \textbf{ data-set } & (mean $\pm$ SD) \\
 \tableHline
    \textbf{ AMS-Grad } & STL 10 & 9.7 $\pm$ 1.0 \\
    \textbf{ AMS-Grad } & SVHN & 1.3 $\pm$ 0.4 \\
    \textbf{ AMS-Grad } & CIFAR 10 & 4.6 $\pm$ 2.2 \\
    \textbf{ AMS-Grad } & CIFAR 100 & 1.1 $\pm$ 1.3 \\
    \textbf{ AMS-Grad } & f MNIST & 1.8 $\pm$ 0.7 \\
 \tableHline
    \textbf{ Adam } & STL 10 & 8.3 $\pm$ 0.8 \\
    \textbf{ Adam } & SVHN & 1.0 $\pm$ 0.6 \\
    \textbf{ Adam } & CIFAR 10 & 3.7 $\pm$ 1.9 \\
    \textbf{ Adam } & CIFAR 100 & 1.6 $\pm$ 1.6 \\
    \textbf{ Adam } & f MNIST & 1.7 $\pm$ 0.8 \\
 \tableHline
    \textbf{ Adam-W } & STL 10 & 8.5 $\pm$ 1.3 \\
    \textbf{ Adam-W } & SVHN & 0.7 $\pm$ 0.4 \\
    \textbf{ Adam-W } & CIFAR 10 & 1.8 $\pm$ 1.3 \\
    \textbf{ Adam-W } & CIFAR 100 & 0.9 $\pm$ 2.9 \\
    \textbf{ Adam-W } & f MNIST & 1.2 $\pm$ 0.5 \\
 \tableHline
    \textbf{ Adamax } & STL 10 & 8.9 $\pm$ 1.1 \\
    \textbf{ Adamax } & SVHN & 1.1 $\pm$ 0.5 \\
    \textbf{ Adamax } & CIFAR 10 & 3.7 $\pm$ 1.5 \\
    \textbf{ Adamax } & CIFAR 100 & 2.3 $\pm$ 2.0 \\
    \textbf{ Adamax } & f MNIST & 1.8 $\pm$ 0.8 \\
 \tableHline
    \textbf{ L1 reg. } & STL 10 & 2.3 $\pm$ 0.6 \\
    \textbf{ L1 reg. } & SVHN & 0.6 $\pm$ 0.8 \\
    \textbf{ L1 reg. } & CIFAR 10 & 1.6 $\pm$ 2.1 \\
    \textbf{ L1 reg. } & CIFAR 100 & 1.8 $\pm$ 2.1 \\
    \textbf{ L1 reg. } & f MNIST & 0.5 $\pm$ 0.7 \\
 \tableHline
    \textbf{ L2 reg. } & STL 10 & 1.8 $\pm$ 1.0 \\
    \textbf{ L2 reg. } & SVHN & 0.6 $\pm$ 0.6 \\
    \textbf{ L2 reg. } & CIFAR 10 & 1.2 $\pm$ 3.1 \\
    \textbf{ L2 reg. } & CIFAR 100 & 1.4 $\pm$ 1.9 \\
    \textbf{ L2 reg. } & f MNIST & 0.5 $\pm$ 0.6 \\
 \tableHline
    \textbf{ Nesterov's } & STL 10 & 0.6 $\pm$ 0.8 \\
    \textbf{ Nesterov's } & SVHN & 0.6 $\pm$ 0.4 \\
    \textbf{ Nesterov's } & CIFAR 10 & 2.7 $\pm$ 1.5 \\
    \textbf{ Nesterov's } & CIFAR 100 & 2.9 $\pm$ 1.8 \\
    \textbf{ Nesterov's } & f MNIST & 0.7 $\pm$ 0.4 \\
 \tableHline
    \textbf{ RMS-prop } & STL 10 & 2.1 $\pm$ 3.3 \\
    \textbf{ RMS-prop } & SVHN & 0.7 $\pm$ 0.3 \\
    \textbf{ RMS-prop } & CIFAR 10 & 1.7 $\pm$ 1.5 \\
    \textbf{ RMS-prop } & CIFAR 100 & -3.9 $\pm$ 3.2 \\
    \textbf{ RMS-prop } & f MNIST & 1.1 $\pm$ 1.2 \\
 \tableHline
    \textbf{ LR decay } & STL 10 & 0.3 $\pm$ 0.3 \\
    \textbf{ LR decay } & SVHN & 1.6 $\pm$ 0.5 \\
    \textbf{ LR decay } & CIFAR 10 & 3.8 $\pm$ 1.0 \\
    \textbf{ LR decay } & CIFAR 100 & 8.5 $\pm$ 1.0 \\
    \textbf{ LR decay } & f MNIST & 0.8 $\pm$ 0.7 \\
 \tableHline
    \textbf{ dropout } & STL 10 & 4.0 $\pm$ 1.2 \\
    \textbf{ dropout } & SVHN & 1.0 $\pm$ 0.5 \\
    \textbf{ dropout } & CIFAR 10 & 1.6 $\pm$ 1.9 \\
    \textbf{ dropout } & CIFAR 100 & 4.4 $\pm$ 2.9 \\
    \textbf{ dropout } & f MNIST & 0.7 $\pm$ 0.5 \\
 \tableHline
    \textbf{ PG } & STL 10 & 4.5 $\pm$ 0.5 \\
    \textbf{ PG } & SVHN & 2.7 $\pm$ 0.8 \\
    \textbf{ PG } & CIFAR 10 & 4.7 $\pm$ 1.8 \\
    \textbf{ PG } & CIFAR 100 & 7.6 $\pm$ 1.8 \\
    \textbf{ PG } & f MNIST & 1.2 $\pm$ 0.8 \\
 \tableHline
\end{tabular}
\caption{\label{table_metaSelection} Data-sets. \upshape{ Characteristics of the data-sets used across experiments. } }
\end{table}

\begin{center}
\begin{table*}[h]
\centering
\begin{tabular}{ m{0.25cm} m{1.5cm} x{1.25} x{1.5} x{1.25} x{1.25} x{1.25} }
\textbf{ } & \textbf{hyperpar.} & \textbf{CIFAR 10} & \textbf{CIFAR 100} & \textbf{fMINST} & \textbf{STL 10} & \textbf{SVHN}  \\
 \tableHline
 \multirow{7}{*}{ \textbf{ \rotatebox{ 90 }{ DenseNet 161 } } } & initial LR & 0.01 & 0.01 & 0.005 & 0.01 & 0.01 \\
      & batch size & 64 & 64 & 128 & 64 & 64 \\
      & pads length & 8 & 8 & 4 & 8 & 8 \\
      & pads type & zeros & zeros & zeros & border & zeros \\
      & h-flip & true & true & true & true & false \\
      & norm mean & 0.8 & 0.2 & none & none & 0.8 \\
      & norm SD & none & 0.8 & 0.1 & 0.2 & none \\
 \tableHline
 \multirow{7}{*}{ \textbf{ \rotatebox{ 90 }{ MobileNet v2 } } } & initial LR & 0.01 & 0.01 & 0.005 & 0.01 & 0.01 \\
      & batch size & 128 & 64 & 128 & 64 & 128 \\
      & pads length & 8 & 8 & 4 & 8 & 8 \\
      & pads type & zeros & zeros & border & zeros & border \\
      & h-flip & true & true & true & true & false \\
      & norm mean & 0.8 & 0.8 & none & 0.8 & 0.8 \\
      & norm SD & none & 0.1 & 0.1 & none & 0.1 \\
 \tableHline
 \multirow{7}{*}{ \textbf{ \rotatebox{ 90 }{ ResNet 18 } } } & initial LR & 0.01 & 0.01 & 0.002 & 0.01 & 0.01 \\
      & batch size & 128 & 64 & 256 & 256 & 128 \\
      & pads length & 8 & 4 & 4 & 8 & 8 \\
      & pads type & zeros & border & reflection & zeros & reflection \\
      & h-flip & true & true & true & true & false \\
      & norm mean & none & 0.2 & none & 0.8 & 0.8 \\
      & norm SD & 0.1 & none & 0.1 & 0.8 & 0.2 \\
 \tableHline
 \multirow{7}{*}{ \textbf{ \rotatebox{ 90 }{ ResNet 50 } } } & initial LR & 0.01 & 0.01 & 0.005 & 0.01 & 0.01 \\
      & batch size & 64 & 64 & 128 & 64 & 64 \\
      & pads length & 8 & 4 & 4 & 8 & 8 \\
      & pads type & zeros & zeros & zeros & zeros & zeros \\
      & h-flip & true & true & true & true & false \\
      & norm mean & 0.2 & 0.2 & none & 0.8 & 0.8 \\
      & norm SD & 0.8 & none & 0.8 & none & 0.2 \\
 \tableHline
 \multirow{7}{*}{ \textbf{ \rotatebox{ 90 }{ ShuffleNet v2 } } } & initial LR & 0.01 & 0.01 & 0.002 & 0.01 & 0.01 \\
      & batch size & 128 & 64 & 256 & 64 & 128 \\
      & pads length & 8 & 4 & 4 & 8 & 4 \\
      & pads type & zeros & zeros & reflection & zeros & zeros \\
      & h-flip & true & true & true & false & false \\
      & norm mean & 0.8 & 0.2 & none & 0.8 & 0.8 \\
      & norm SD & 0.8 & none & 0.2 & 0.8 & 0.2 \\
 \tableHline
\end{tabular}
\vspace*{ \tableMarginBottom }
\caption{\label{table_metaparameters} Tuned hyper-parameters. \upshape{ Training hyper-parameters were fine-tuned to optimize performance for each individual baseline architecture $\times$ data-set. The table shows the best performing hyper-parameters found for each case. } }
\end{table*}
\end{center}

\begin{figure*}[t]
\centering
\includegraphics[ trim = 0cm 0cm 0cm 0cm, clip, width = 17cm ]{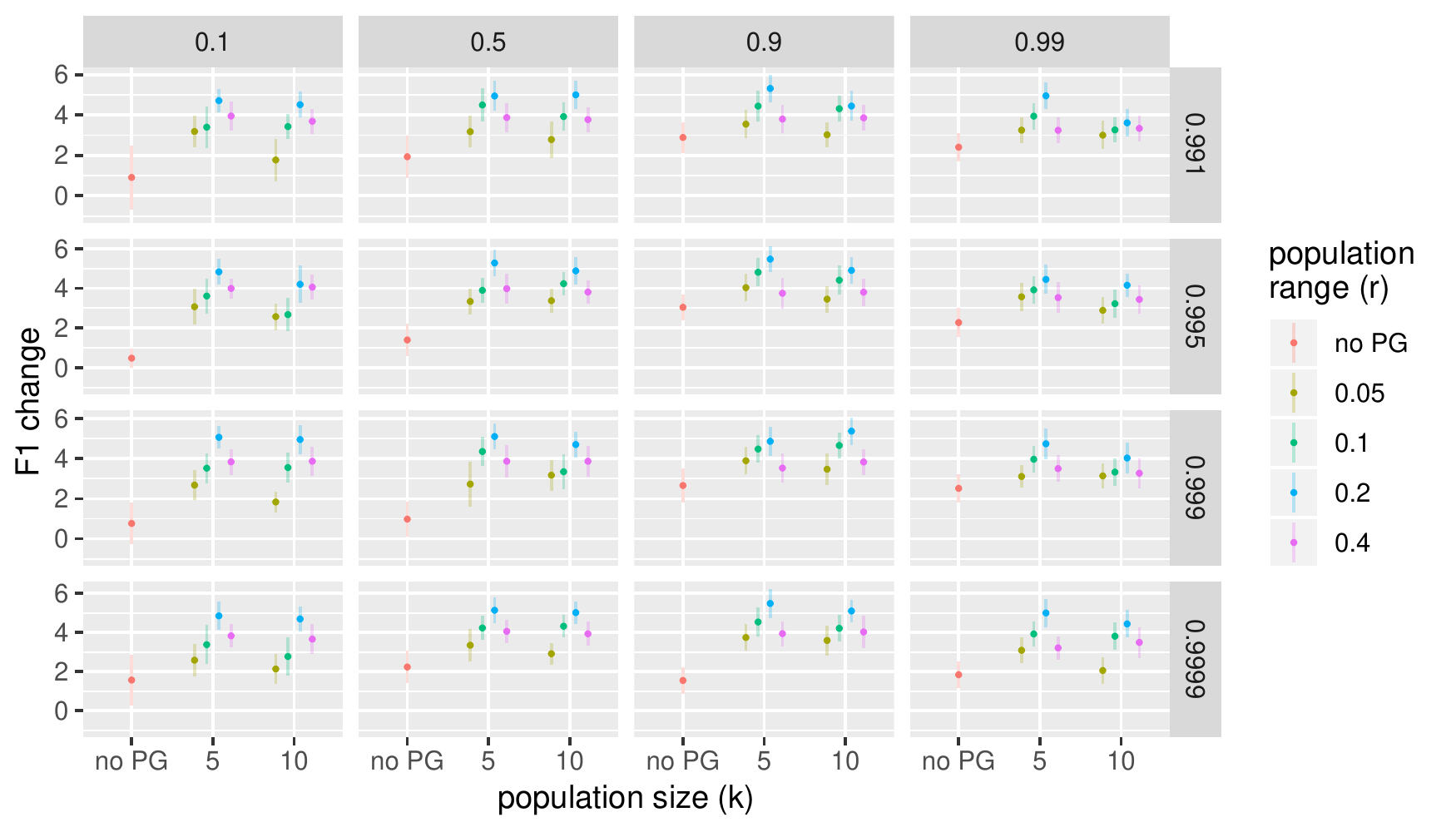}
\caption{ \label{fig_sim} Combining with other methods. The figure shows performance improvement (y-axis, testing set F1) of baseline ResNet18 on CIFAR 10 when it combines PG with Adamax. This is shown separately for each possible combination of Adamax's parameters (beta 1 in columns of panels, beta 1 in columns of panels), and PG's parameters (population size 's' in x-axis, population range 'r' in y-axis). In each case, averages (dots) and standard deviations (bars) are calculated across the last 10 epochs of training. Official training/testing splits of samples used. }
\end{figure*}




\printbibliography

\begin{IEEEbiography}{Michael Shell}
Biography text here.
\end{IEEEbiography}

\end{document}